\documentclass[runningheads]{llncs}
\usepackage{graphicx}
\usepackage{subcaption}
\usepackage[colorlinks=false, hidelinks,breaklinks=true]{hyperref}

\usepackage{marvosym}

\usepackage[T1]{fontenc}
\usepackage{graphicx}
\begin{document}
\sloppy
\title{Evaluating Compliance with Visualization Guidelines in Diagrams for Scientific Publications Using Large Vision Language Models}
\titlerunning{Evaluating Visualization Guideline Compliance in Diagrams Using VLMs}
%
\author{Johannes Rückert\inst{1}\thanks{these authors contributed equally to the work}\orcidID{0000-0002-5038-5899}\textsuperscript{(\Letter)} \and
Louise Bloch\inst{1,2,3}$^\star$\orcidID{0000-0001-7540-4980} \and
Christoph M. Friedrich\inst{1,2}\orcidID{0000-0002-3315-7536}}
%
\authorrunning{J. Rückert et al.}
%
\institute{Department of Computer Science, University of Applied Sciences and Arts Dortmund, Emil-Figge-Str. 42, 44227 Dortmund, Germany \email{\{johannes.rueckert,louise.bloch,christoph.friedrich\}@fh-dortmund.de}\and
Institute for Medical Informatics, Biometry and Epidemiology (IMIBE), 45147 Essen, Germany \and
Institute for Artificial Intelligence in Medicine (IKIM), University Hospital Essen, 45147 Essen, Germany}

\maketitle              
\begin{abstract}
Diagrams are widely used to visualize data in publications. The research field of data visualization deals with defining principles and guidelines for the creation and use of these diagrams, which are often not known or adhered to by researchers, leading to misinformation caused by providing inaccurate or incomplete information.

In this work, large Vision Language Models (VLMs) are used to analyze diagrams in order to identify potential problems in regards to selected data visualization principles and guidelines. To determine the suitability of VLMs for these tasks, five open source VLMs and five prompting strategies are compared using a set of questions derived from selected data visualization guidelines.

The results show that the employed VLMs work well to accurately analyze diagram types ($F_1$-score 82.49~\%), 3D effects ($F_1$-score 98.55~\%), axes labels ($F_1$-score 76.74~\%), lines (RMSE 1.16), colors (RMSE 1.60) and legends ($F_1$-score 96.64~\%, RMSE 0.70), while they cannot reliably provide feedback about the image quality ($F_1$-score 0.74~\%) and tick marks/labels ($F_1$-score 46.13~\%). Among the employed VLMs, Qwen2.5VL performs best, and the summarizing prompting strategy performs best for most of the experimental questions.

It is shown that VLMs can be used to automatically identify a number of potential issues in diagrams, such as missing axes labels, missing legends, and unnecessary 3D effects. The approach laid out in this work can be extended for further aspects of data visualization.

\keywords{Document Analysis Systems \and Chart Understanding \and Visualization Guidelines  \and Vision Language Models.}
\end{abstract}

\section{Introduction}
Diagrams can help researchers communicate their results more effectively. Previous research~\cite{Tahamtan2016FactorsAffectingCitations} found that the use of figures increases the citation counts of scientific publications across disciplines. This suggests that figures make information easier to understand and more convincing. Data visualization is a field of research focusing on how to represent data for clear, effective communication. However, previous research~\cite{Parson2022UnderstandingDesignPractice} found that many practitioners are not aware of these principles and guidelines. Additionally, there are many pitfalls in data visualization which may lead to misinformation caused by providing inaccurate or incomplete information~\cite{Bresciani2015PitfallsofVisualRepresentations,Nguyen2021DataVisualizationPitfalls}. Visualization guidelines~\cite{Jambor2024FromZeroToHeroChecklist,Tufte2001visualDisplayOfQuantitativeInformation} have been developed, providing detailed explanations of how to best present data. These detailed explanations are suitable for experts in data visualization. However, diagrams are created by an interdisciplinary community, making it complicated and time-consuming to understand and follow detailed guidelines. More recent publications provide tools such as checklists~\cite{Jambor2024FromZeroToHeroChecklist} or visualization linters~\cite{Mcnutt2018LintingFV} to make such guidelines accessible to a wider audience. Another approach is to develop tools that assess a diagram's quality and provide feedback that can be useful for both diagram creators and readers. In \cite{Shukla2008QualityAssessmentOfCharts}, a diagram evaluation system based on classical image and text processing was evaluated. The resulting quality score includes the spatial proximity between the figure and the text reference, the completeness of the identified labels, the image contrast, and the consistency between reference texts, captions, and labels. The score shows a moderate to high correlation with a human evaluation on $15$ figures. 

It is assumed that such tasks can also be performed using recent Large Language Models (LLMs). In~\cite{kim2024goodchatgptgivingadvice} the human performance in advising data visualization is compared to ChatGPT v3.5. Human responses to $119$ questions in the visualization forum VisGuides~\cite{Diehl2018VisGuidesForum} were compared to those of ChatGPT. Two experts rated the responses according to six criteria (coverage, topicality, breadth, clarity, depth, and actionability). The results show that ChatGPT and human responses perform comparably, especially in clarity. Humans showed higher variability. Twelve participants discussed visualization designs with human experts and ChatGPT. All participants preferred expert discussions for their fluidity and tailored recommendations. For ChatGPT, broad knowledge, neutrality and its ability for quick brainstorming were mentioned positively.

A system for evaluating diagrams and providing feedback to novice designers was introduced in~\cite{Shin2024visualizationaryautomatingdesignfeedback}. Preprocessing steps were combined with LLMs to provide interpretations and suggestions. Feedback covers areas such as title suggestions, diagram type choice, removal of chart junk, and use of accessible, distinguishable colors. The preprocessing steps include plot-to-table conversion with DePlot~\cite{Liu2023deplot}, YOLOR~\cite{Wang2021YOLOR} to detect chart junk, PyTesseract-OCR to recognize text in diagrams, and Scanner Deeply~\cite{Shin2023ScannerDeeply} as a virtual eye tracker. The preprocessed results as well as visualization guidelines were used to generate interpretations and suggestions using ChatGPT. The results of a longitudinal study, including six novice, four intermediate, and three expert visualization designers, show that the tool can help advanced designers refine their diagrams. 

This research adopts a similar approach but relies on open source Vision-Language Models (VLMs) rather than employing complex image and text processing techniques. Unlike traditional deep neural networks, VLMs are trained on vast amounts of image-text pairs, allowing them to perform zero-shot predictions for various tasks~\cite{Zhang2024VLM}. VLMs have been successfully used for tasks such as chart question answering~\cite{Masry2022ChartqaDataset}, chart summarization~\cite{Kantharaj2022ChartSummarization}, and chart-to-data extraction~\cite{Liu2023deplot}. To the best of our knowledge, no previous research used open source VLMs to check the compliance of scientifically published diagrams with visualization guidelines. The contributions of this research are as follows:
\begin{itemize}
    \item Examine the compliance with visualization guidelines  present in diagrams featured in scientific publications 
    \item Evaluate the performance of various open source VLM models in assessing the compliance of diagrams with visualization guidelines
\end{itemize}
\section{Materials and Methods}

\begin{figure}[t!]
\centering
\includegraphics[width=0.78\textwidth]{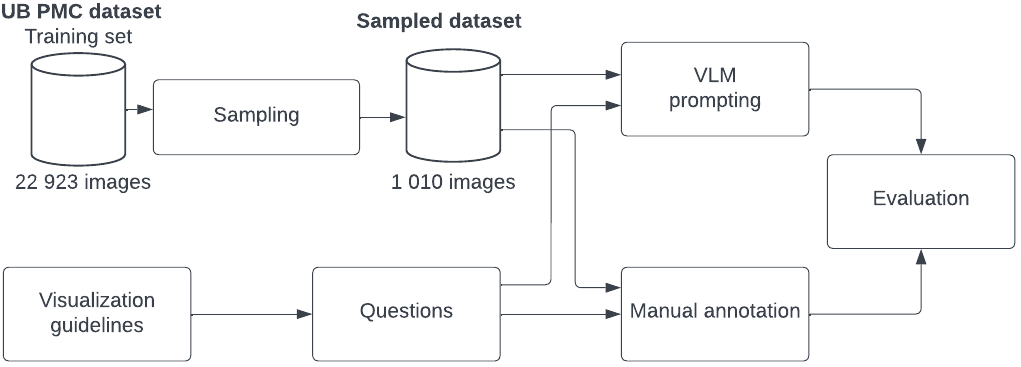}
\caption{Overview of the general workflow.} \label{fig:workflow}
\end{figure}

An overview of the workflow is given in Fig.~\ref{fig:workflow}. A small subset of images was sampled from the University at Buffalo PubMedCentral (UB PMC) chart dataset (Section~\ref{SEC:Dataset}). A set of questions was derived from visualization guidelines (Section~\ref{SEC:VisualizationGuidelines}) to assess the VLM's ability to identify violations and improve diagram quality. VLM prompts were designed for these questions, aiming for answers which can be automatically parsed. Ground truth annotations were created based on existing annotations from the UB PMC dataset and a manual labeling process.
Five open source VLMs (Section~\ref{SEC:VLMs}) are compared for diagram type classification. For the best-performing model, five prompting strategies (Section~\ref{SEC:Prompting}) are compared. Details about the sampled dataset, the VLM prompting pipeline, the evaluation scripts, and supplementary materials are available in a git repository\footnote{\url{https://gitlab.com/ippolis_wp3/icdar2025-diagram-analysis-vlm}, accessed 2025-06-18}.


%

\subsection{Dataset}
\label{SEC:Dataset}
The experiments were performed using the UB PMC dataset~\cite{Davila2022CHARTInfographics}, which is based on images of diagrams extracted from papers sampled from the Open Access section of the PMC archive and has been published for the 2022 edition of the Challenge on Harvesting Raw Tables from Infographics (CHART-Infographics). For all diagrams, manual annotations of the diagram type are provided. In addition, the texts and text positions, as well as information about the positions of legends, axes and other diagram elements for certain types of diagrams are given. 
Containing 22\,923 images in the training set and 13\,260 in the test set, the dataset covers 15 diagram types (area, line, Manhattan, scatter, scatter-line, pie, vertical and horizontal box, vertical and horizontal bar, vertical and horizontal interval, map, heatmap, surface, and Venn diagrams). For the dataset used in our experiments, $1\,010$ images were randomly sampled from the training set roughly corresponding to the distribution of diagram types in the original dataset. A minimal number of $30$ images for each type was ensured. Horizontal interval and map diagrams were excluded from the dataset. The images in the sampled dataset have an average width of 641~px and an average height of 458~px. The ground truth labels for our experiments were derived from the labels provided in the UB PMC dataset where possible, and manually annotated by two annotators otherwise.
The original and sampled datasets are summarized in Table~\ref{Table:OriginalDataset}. 
\begin{table}[t!]
    \centering
    \caption{Distribution of the diagram types of the UB PMC dataset and our sampled dataset, which is selected from the training set. Horizontal and vertical diagram types, marked with an asterisk, were consolidated for further analysis.}
    \label{Table:OriginalDataset}
    \begin{tabular}{|l|r|r|r||r|}
    \hline
         Diagram Type & Training set & Test set& $\sum$ & Sampled set \\\hline
Area chart & 172 & 136 & 308 & 30\\
Line chart&  10\,556 & 3\,400&13\,956& 200\\
Manhattan chart & 176 & 80&256& 30\\
Scatter plot& 1\,350 & 1\,247&2\,597& 100\\
Scatter-line plot & 1\,818 & 1\,628&3\,446& 100\\
Pie chart& 242 & 191&433& 30\\
Vertical box chart & 763 & 775&1\,538& 100*\\
Horizontal bar chart & 787  & 634&1\,421& 100*\\
Vertical bar chart & 5\,454 & 3\,745&9\,199& 200*\\
Horizontal interval chart & 156  & 430&586& 0*\\
Vertical interval chart & 489  & 182 & 671& 30*\\
Map & 533  & 373&906& 0\\
Heatmap & 197  & 180&377& 30\\
Surface & 155  & 128&283& 30\\
Venn & 75  & 131&206& 30\\\hline
$\sum$& 22\,923 &13\,260&36\,183& 1\,010\\\hline
    \end{tabular}
\end{table}



\subsection{Visualization Guidelines}\label{SEC:VisualizationGuidelines}

Visualization guidelines~\cite{Bresciani2015PitfallsofVisualRepresentations,Franzblau2012VisualizationGuidelines,Jambor2024FromZeroToHeroChecklist,Nguyen2021DataVisualizationPitfalls,Park2022PrinciplesOfPresentingStatisticalResultsFigures,Schriger2001VisualizationGuidelines,Tufte2001visualDisplayOfQuantitativeInformation} aim to establish best practices for creating effective and accurate data diagrams.
The visualization guidelines selected in this research include the avoidance of 3D diagrams, especially when one axis lacks information,~\cite{Franzblau2012VisualizationGuidelines,Schriger2001VisualizationGuidelines} \cite[p. 96]{Tufte2001visualDisplayOfQuantitativeInformation} and the avoidance of pie diagrams due to difficult angle comparisons \cite[p. 178]{Tufte2001visualDisplayOfQuantitativeInformation}. Additionally, color should be used sparingly~\cite{Jambor2024FromZeroToHeroChecklist,Nguyen2021DataVisualizationPitfalls} \cite[p. 96]{Tufte2001visualDisplayOfQuantitativeInformation}, and diagrams with axes should have labels~\cite{Jambor2024FromZeroToHeroChecklist,Mcnutt2018LintingFV,Park2022PrinciplesOfPresentingStatisticalResultsFigures,Schriger2001VisualizationGuidelines}, as well as tick marks and tick labels~\cite{Jambor2024FromZeroToHeroChecklist,Park2022PrinciplesOfPresentingStatisticalResultsFigures} for all axes. Legends should be used to explain all diagram elements~\cite{Schriger2001VisualizationGuidelines}. If line diagrams are used, lines should be used sparingly. 
Overall, the diagrams should not contain any compression artifacts.
From these guidelines, 13 questions were derived for the VLMs:

    \begin{enumerate}
        \item Which of the following categories best describes the diagram type? surface, box, pie, scatter-line, area, scatter, bar, interval, venn, line, heatmap, manhattan.
        \item Does the diagram visually appear to have a 3D effect?
        \item Do all the axes have labels?
        \item Does the horizontal axis contain an axis label?
        \item Does the vertical axis contain an axis label?
        \item Do all axes have tick marks and tick labels?
        \item Does the horizontal axis have tick marks and tick labels?
        \item Does the vertical axis have tick marks and tick labels?
        \item How many lines does the diagram contain? Do not count the axes.
       \item How many colors are used in the diagram? Do not count black as a color.
         \item Does the diagram contain a legend?
         \item How many groups are used in the legend of the diagram?
       \item Does the image contain any compression artifacts (such as visible block structures and halos around edges or general loss of sharpness and oscillations around high-contrast edges)?
   \end{enumerate}



\subsection{Vision-Language Models}
\label{SEC:VLMs}

Due to their abilities in text generation and comprehension, LLMs have recently attracted many researchers~\cite{Kasneci2023OpportunitiesAndChallengesChatGPTForEducation}. The attention mechanism~\cite{Vaswani2017AttentionMechanism} and transformer architecture~\cite{Devlin2019BERT} enhanced the capacity of LLMs to handle long-range dependencies in texts. Originally, LLMs deal with text inputs only, but there is a large number of multimodal use cases combining texts with images, videos or speech. In recent years, a range of large VLMs~\cite{Alayrac2022flamingo,Bai2023qwenvlversatilevisionlanguagemodel,hong2024cogvlm2,Su2020VLBERT,Wang2022simVLM} have been developed. In this research, four general-purpose VLMs and one chart-specific VLM were compared in terms of their suitability for answering questions about diagrams.

CogVLM2~\cite{hong2024cogvlm2} consists of four components: a Vision Transformer (ViT) \cite{Dosovitskiy2021VIT}, a vision-language adapter, a language model (Llama3-8B), and a visual expert module~\cite{wang2024cogvlmvisualexpertpretrained}.

InternVL2.5~\cite{chen2025expanding} is a Multimodal Large Language Model (MLLM) and uses a similar structure as CogVLM2: A ViT (InternViT) is combined with an LLM (InternLM2.5) via an adapter.

Janus-Pro~\cite{chen2025januspro} is a multimodal framework decoupling encoding and generation into separate pathways. An auto-regressive transformer acts as the adapter between separate encoders and decoders for different input and output modalities.  

The most recent VLM is Qwen2.5VL~\cite{Bai2023qwenvlversatilevisionlanguagemodel,bai2025qwen25vltechnicalreport}. It consists of three main components which are the LLM (Qwen2.5~\cite{qwen2025qwen25technicalreport}), the visual encoder and the vision-language adapter. The ViT architecture used as the visual encoder is re-designed compared to previous QwenVL iterations to better handle images of varying sizes, and trained from scratch. 
In addition to the 7B variant used for most experiments, we also tested the 72B Qwen2.5VL model.

The diagram-specific VLM is ChartInstruct~\cite{masry2024chartinstruct}.
ChartInstruct utilizes the Large Language and Vision Assistant (LLaVA)~\cite{liu2023visualinstructiontuning} architecture, which uses Contrastive Language-Image Pre-Training (CLIP)~\cite{clip} for visual encoding, ChartLlama~\cite{han2023chartllamamultimodalllmchart} for language generation, and an adapter, similar to QwenVL. ChartLlama is pre-trained on a chart dataset specifically created for this purpose. 

Table~\ref{Table:VLMInfo} gives an overview of the VLMs.
\begin{table}[t!]
    \centering
    \caption{Overview of the LLMs used in the VLMs and their parameter counts, license and accepted image resolution. Image resolutions marked with an asterisk are dynamic and denote the maximum supported resolution.}
    \label{Table:VLMInfo}
    \begin{tabular}{|l|l|l|r|r|}
    \hline
         Model & License & LLM & \# Parameters & Image resolution \\
         \hline\hline
         CogVLM2 & CogVLM & Llama3 & 8B & 1344$\times$1344 \\
         InternVL2.5 & MIT & InternLM2.5 & 7B & 448$\times$448 \\
         
         Janus-Pro & DeepSeek & DeepSeek-LLM & 7B & 384$\times$384 \\
         Qwen2.5VL & Apache 2.0 & Qwen2.5LM & 7B & 3584$\times$3584* \\
         ChartInstruct & GPLv3 & Llama2 & 7B & 224$\times$224 \\
         \hline
    \end{tabular}
\end{table}
The VLMs were run with default settings, using beam search for the text generation. After the initial VLM comparison, a sampling text generation configuration was tested.

\subsection{Prompting}
\label{SEC:Prompting}

For the general purpose VLMs the system prompts were initialized with ``You are a helpful assistant specializing in scientific visualization. Your task is to help researchers assess and improve plots in their scientific papers.'' followed by a specification of the expected response to facilitate automatic processing:

\begin{itemize}
    \item \textbf{Yes/no questions:} ``Please answer the question with a single "yes" or "no".''
    \item \textbf{Numeric questions:} ``Please answer the question with a single whole number (e.g. 1, 2, or 3).''
    \item \textbf{Diagram type:} ``Please answer the question with one of the given options only.''
\end{itemize}

The chart-specific model does not support system prompts and was prompted with the question followed by one of the above answer type specifications. 
The generated answers were post-processed to try and extract the relevant information if a model did not comply with the expected answer type specification:

\begin{itemize}
    \item \textbf{Yes/no questions:} The lower-cased answer is searched for occurrences of ``yes''. If none are found, it is searched for occurrences of ``no''.
    \item \textbf{Numeric questions:} The answer is searched for numbers. If more than one is found, the highest number is used. If no numbers were found, numeric words (from ``one'' to ``fifteen'') are searched.
    \item \textbf{Diagram type:} The answer is searched for the diagram types.
\end{itemize}


Further, we compared the performance of five prompting strategies.
\begin{itemize}
    \item \textbf{Individual questions:} Each question is asked in a separate conversation context.
    \item \textbf{Context:} All questions are asked in the same conversation context.
    \item \textbf{Elaborate:} Each question is asked in a separate context, but after the initial answer, the model is asked ``Are you sure? Please elaborate on your answer.'' before being asked for a final answer.
    \item \textbf{Summary:} Each question is asked in a separate conversation context, but initially without specifying the answer type. Afterwards, the VLM is asked to summarize its answer using the answer type specification.
    \item \textbf{Introduction questions:} Questions are split into groups, which are asked in the same conversation context. Before asking each set of questions, the VLM is prompted to describe the image regarding relevant aspects, e.g., ``Please describe the axes, tick marks and tick labels of the diagram.''
\end{itemize}

In addition to these zero-shot strategies, a few-shot approach was implemented for the individual question strategy.


\section{Experiments and Results}
The primary objective is to evaluate the ability of VLMs to detect violations of visualization guidelines in scientific diagrams. This evaluation compares four general-purpose VLMs with one chart-specific VLM to determine whether the latter outperforms the former in diagram-specific tasks. In addition, five prompting strategies are tested for the best-performing VLM. The secondary focus is identifying violated visualization guidelines in scientific diagrams.

\subsection{Diagram Types}
The VLM's suitability for diagram analysis is initially investigated by a diagram type classification (Section~\ref{SEC:VisualizationGuidelines} question 1). These experiments serve two purposes. First, some subsequent experiments apply to specific diagram types; for instance, axis-related analyses are irrelevant for pie diagrams. Here, the analysis functions as a preprocessing step. Second, certain visualization guidelines advise against using pie charts. Since the orientation of the diagrams did not affect the subsequent experiments, the vertical and horizontal diagram types were merged.

\begin{table}[t!]
    \centering
    \caption{Findings on the recognition of diagram types (Section~\ref{SEC:VisualizationGuidelines} question 1) in $1\,010$ images. The ChartInstruct model produced invalid results for four images ($0.40~\%$). Best results are highlighted in bold. No information rate: $29.70~\%$.}
    \label{Table:ResultsDiagramTypes}
    \begin{tabular}{|l|r|r|r|r|r|}
    \hline
         Model & Precision & Recall &  $F_1$-score & Accuracy & Balanced accuracy\\
         \hline\hline
         CogVLM2&$71.98~\%$&$65.44~\%$&$61.74~\%$&$74.65~\%$&$65.44~\%$\\
         InternVL2.5&$77.60~\%$&$67.99~\%$&$66.50~\%$&$76.53~\%$&$67.99~\%$\\
         
         Janus-Pro&$76.87~\%$&$62.57~\%$&$60.89~\%$&$68.91~\%$&$62.57~\%$\\
         Qwen2.5VL&$\mathbf{92.47~\%}$&$\mathbf{80.76~\%}$&$\mathbf{82.49~\%}$&$\mathbf{84.75~\%}$&$\mathbf{80.76~\%}$\\
         ChartInstruct&$2.72~\%$&$9.56~\%$&$3.64~\%$&$11.38~\%$&$10.36~\%$\\
         \hline
         \end{tabular}
         \end{table}
Table~\ref{Table:ResultsDiagramTypes} summarizes the performances (macro-averaged precision, macro-averaged recall, macro-averaged $F_1$-score, accuracy, and balanced accuracy) of five VLMs. Performances for the 72B model and the sampling approach, as well as matrix plots illustrating the cross-tables of model predictions and manual labels are included in the supplementary materials. The ChartInstruct model achieves the lowest $F_1$-score ($3.64~\%$). This model classifies $69.41~\%$ of the diagrams as scatter-line plots and $26.44~\%$ as scatter plots. Invalid answers were produced for four images ($0.40~\%$). The best results were achieved for the Qwen2.5VL ($F_1$-score: $82.49~\%$).
For this model, it was tested whether a different text generation configuration improves the results. The default beam search was replaced with the sampling strategy. This approach slightly decreases the performance ($F_1$-score: $81.07~\%$). Based on these results, the remaining experiments will use the base Qwen2.5VL model. The performance of a larger 72B version of the Qwen2.5VL model outperforms the results reached by the 7B model ($F_1$-score: $85.93~\%$). Due to the higher utilization of resources and the inference time, the 7B version was used to test prompting strategies.

\subsection{3D Diagrams}
   \begin{table}[t!]
    \centering
    \caption{Results on the identification of 3D effects in $1\,010$ diagrams (Section~\ref{SEC:VisualizationGuidelines} question 2). None of the strategies produced invalid results. The best results are highlighted in bold. No information rate: $96.53~\%$.}
    \label{Table:Results3DPlots}
    \begin{tabular}{|l|r|r|r|r|r|}
    \hline
         Prompting strategy & Precision & Recall &  $F_1$-score & Accuracy &Balanced accuracy\\
         \hline\hline
         Individual question&$\mathbf{100.00~\%}$&$\mathbf{97.14~\%}$&$\mathbf{98.55~\%}$&$\mathbf{99.90~\%}$&$\mathbf{98.57~\%}$\\
         Context&$\mathbf{100.00~\%}$&$94.29~\%$&$97.06~\%$&$99.80~\%$&$97.14~\%$\\
         Elaborate&$\mathbf{100.00~\%}$&$\mathbf{97.14~\%}$&$\mathbf{98.55~\%}$&$\mathbf{99.90~\%}$&$\mathbf{98.57~\%}$\\
        Summary&$\mathbf{100.00~\%}$&$94.29~\%$&$97.06~\%$&$99.80~\%$&$97.14~\%$\\
       Introduction question&$\mathbf{100.00~\%}$&$94.29~\%$&$97.06~\%$&$99.80~\%$&$97.14~\%$\\
       \hline

         \end{tabular}
         \end{table}
Some visualization guidelines~\cite{Franzblau2012VisualizationGuidelines,Schriger2001VisualizationGuidelines}, \cite[p. 96]{Tufte2001visualDisplayOfQuantitativeInformation} discourage from using 3D diagrams, as they can distract the reader (Section~\ref{SEC:VisualizationGuidelines} question 2). The necessity of the 3D effect can be inferred from the diagram type. For example, bar and pie charts should avoid it, unlike surface plots. The results are summarized in Table~\ref{Table:Results3DPlots}. Results for the 72B model and the few-shot approach are included in the supplementary materials. 3D effects were found in $35$ ($3.47~\%$) diagrams (no information rate: $96.53~\%$). These include 30 surface plots ($85.71~\%$) and five pie charts ($14.29~\%$). All strategies outperformed the no information rate. The best $F_1$-score was $98.55~\%$ (individual question, and elaborate). All strategies reached perfect precision. The few-shot approach achieved perfect classification, while the larger Qwen2.5VL-72B did not yield any improvements.

\subsection{Axes}
\begin{table}[t!]
    \centering
    \caption{Results on the identification, and localization of axes labels (Section~\ref{SEC:VisualizationGuidelines} question 3-5). None of the strategies produced invalid results. The best results are highlighted in bold.}
    \label{Table:ResultsAxisLabels}
    \begin{tabular}{|l|r|r|r|r|r|}
    \hline
         Prompting strategy & Precision & Recall &  $F_1$-score & Accuracy &Balanced accuracy\\
          \hline\hline
         \multicolumn{6}{|c|}{All axes labels (question 3) ($n=950$, No information rate: $70.21~\%$)}\\
         \hline
         Individual question&$97.78~\%$&$15.55~\%$&$26.83~\%$&$74.74~\%$&$57.70~\%$\\
         Context&$\mathbf{100.00~\%}$&$13.07~\%$&$23.13~\%$&$74.11~\%$&$56.54~\%$\\
         Elaborate&$94.34~\%$&$17.67~\%$&$29.76~\%$&$75.16~\%$&$58.61~\%$\\
         Summary&$84.98~\%$&$\mathbf{69.96~\%}$&$\mathbf{76.74~\%}$&$\mathbf{87.37~\%}$&$\mathbf{82.36~\%}$\\
         Introduction question&$93.22~\%$&$19.43~\%$&$32.16~\%$&$75.58~\%$&$59.42~\%$\\
         \hline
         \multicolumn{6}{|c|}{Horizontal axes labels (question 4) ($n=920$, No information rate: $74.57~\%$)}\\
         \hline
         Individual question&$90.27~\%$&$87.18~\%$&$88.70~\%$&$94.35~\%$&$91.99~\%$\\
         Context&$90.48~\%$&$8.12~\%$&$14.90~\%$&$76.41~\%$&$53.91~\%$\\
         Elaborate&$90.67~\%$&$87.18~\%$&$88.89~\%$&$94.46~\%$&$92.06~\%$\\
         Summary&$88.45~\%$&$\mathbf{94.87~\%}$&$\mathbf{91.55~\%}$&$\mathbf{95.54~\%}$&$\mathbf{95.32~\%}$\\    
         Introduction question&$\mathbf{92.44~\%}$&$47.01~\%$&$62.32~\%$&$85.54~\%$&$72.85~\%$\\
            \hline     
         \multicolumn{6}{|c|}{Vertical axes labels (question 5) ($n=920$, No information rate: $86.63~\%$)}\\
         \hline
         Individual question&$84.35~\%$&$78.86~\%$&$\mathbf{81.51~\%}$&$\mathbf{95.22~\%}$&$88.30~\%$\\
         Context&$\mathbf{100.00~\%}$&$17.89~\%$&$30.34~\%$&$89.02~\%$&$58.94~\%$\\
         Elaborate&$83.62~\%$&$78.86~\%$&$81.17~\%$&$95.11~\%$&$88.24~\%$\\
       Summary&$76.26~\%$&$\mathbf{86.18~\%}$&$80.92~\%$&$94.57~\%$&$\mathbf{91.02~\%}$\\
         Introduction Question&$88.89~\%$&$32.52~\%$&$47.62~\%$&$90.43~\%$&$65.95~\%$\\
         \hline
         \end{tabular}
         \end{table}

The axes and proper labeling are crucial for diagram orientation and understanding. This research considered two visualization guidelines: the completeness of axes labels, tick marks, and tick labels. 

The results considering the axes labels (Section~\ref{SEC:VisualizationGuidelines} question 3-5) are performed on $950$ images and are summarized in Table~\ref{Table:ResultsAxisLabels}. Results for the 72B model and the few-shot approach are included in the supplementary materials. As pie and Venn diagrams lack axes, $60$ images were excluded. Complete labels (all axes) were identified for 667 diagrams (no information rate: $70.21~\%$). All prompting strategies outperformed this rate. The summary strategy reached the best $F_1$-score ($76.74~\%$). All remaining strategies suffered from low recall ($13.07~\%$ - $19.43~\%$). Thus, many diagrams without axis labels were not detected by these strategies. Compared to the individual question approach, the few-shot method improved the $F_1$-score to $66.79~\%$, driven by higher recall. The larger Qwen2.5VL-72B reached an $F_1$-score of $78.17~\%$ (individual question). The summary strategy is expected to further improve these results.

Further investigations on $920$ images (previous set without surface plots) evaluated the VLM's ability to distinguish between missing horizontal and vertical axis labels. For 234 ($25.43~\%$) images, a missing horizontal axis was identified (no information rate: $74.57~\%$). All prompting strategies outperformed this rate. Similarly to the previous results, the summary strategy performed best ($F_1$-score: $91.55~\%$). The context strategy, relying on previous answers, showed low recall ($8.12~\%$), likely due to the VLM repeating earlier mistakes. The few-shot approach achieved an $F_1$-score of $31.98~\%$ and the Qwen2.5VL-72B model of $93.86~\%$. Vertical axis labels are less frequently missing, occurring in 123 ($13.37~\%$) diagrams (no information rate: $86.63~\%$). All strategies exceeded the no information rate. The best-performing strategy (individual question) reached an $F_1$-score of $81.51~\%$. A poor recall ($17.89~\%$) was reached for the context strategy. The results of the Qwen2.5VL-72B model outperformed these results ($F_1$-score: $91.54~\%$).

\begin{table}[t!]
    \centering
    \caption{Results on the identification, and localization of tick marks and tick labels (Section~\ref{SEC:VisualizationGuidelines} question 6-8). None of the strategies produced invalid results. The best results are highlighted in bold.}
    \label{Table:ResultsTickMarks}
    \begin{tabular}{|l|r|r|r|r|r|}
    \hline
         Prompting strategy & Precision & Recall &  $F_1$-score & Accuracy &Balanced accuracy\\
         \hline\hline
         \multicolumn{6}{|c|}{Tick labels and marks (question 6) ($n=950$, No information rate: $77.05~\%$)}\\
         \hline
         Individual question&$56.76~\%$&$9.63~\%$&$16.47~\%$&$77.58~\%$&$53.72~\%$\\
         Context&$26.86~\%$&$34.86~\%$&$30.34~\%$&$63.26~\%$&$53.29~\%$\\
         Elaborate&$57.50~\%$&$10.55~\%$&$17.83~\%$&$77.68~\%$&$54.11~\%$\\
          Summary&$36.91~\%$&$\mathbf{61.47~\%}$&$\mathbf{46.13~\%}$&$67.05~\%$&$\mathbf{65.09~\%}$\\
         Introduction question&$\mathbf{72.22~\%}$&$5.96~\%$&$11.02~\%$&$\mathbf{77.89~\%}$&$52.64~\%$\\
        \hline
         \multicolumn{6}{|c|}{Horizontal tick labels and marks (question 7) ($n=920$, No information rate: $81.63~\%$)}\\
         \hline
        Individual question&$43.00~\%$&$25.44~\%$&$31.97~\%$&$80.11~\%$&$58.93~\%$\\
         Context&$19.08~\%$&$\mathbf{29.59~\%}$&$23.20~\%$&$64.02~\%$&$50.68~\%$\\
         Elaborate&$42.72~\%$&$26.04~\%$&$\mathbf{32.35~\%}$&$80.00~\%$&$\mathbf{59.09~\%}$\\
        Summary&$50.67~\%$&$22.49~\%$&$31.15~\%$&$81.74~\%$&$58.78~\%$\\
         Introduction question&$\mathbf{72.22~\%}$&$7.69~\%$&$13.90~\%$&$\mathbf{82.50~\%}$&$53.51~\%$\\
        
         \hline
         \multicolumn{6}{|c|}{Vertical tick labels and marks (question 8) ($n=920$, No information rate: $86.20~\%$)}\\
         \hline
         Individual question&$52.31~\%$&$26.77~\%$&$35.42~\%$&$86.52~\%$&$61.43~\%$\\
         Context&$18.21~\%$&$\mathbf{40.16~\%}$&$25.06~\%$&$66.85~\%$&$55.64~\%$\\
         Elaborate&$52.38~\%$&$25.98~\%$&$34.74~\%$&$86.52~\%$&$61.10~\%$\\
        Summary&$56.06~\%$&$29.13~\%$&$\mathbf{38.34~\%}$&$87.07~\%$&$\mathbf{62.74~\%}$\\
         Introduction question&$\mathbf{80.00~\%}$&$9.45~\%$&$16.90~\%$&$\mathbf{87.17~\%}$&$54.54~\%$\\
         
         \hline
         \end{tabular}
         \end{table}

Table~\ref{Table:ResultsTickMarks} summarizes the VLM's ability to detect missing tick marks and labels (Section~\ref{SEC:VisualizationGuidelines} question 6-8). Results for the 72B model and the few-shot approach are included in the supplementary materials. 218 images ($22.95~\%$) have missing tick marks or labels (all axes) (no information rate: $77.05~\%$). Three strategies slightly outperformed the no information rate. The introduction question strategy, which reached the best accuracy ($77.89~\%$), achieved a poor recall ($5.96~\%$). Two additional strategies (elaborate, and introduction question) reached a recall smaller than $15.00~\%$. The context and summary strategies suffered from poor precision smaller than $40.00~\%$. The best $F_1$-score was $46.13~\%$ (summary strategy). The results indicate that the VLM struggles to identify tick marks and labels in the diagrams. The few-shot strategy ($F_1$-score: $40.14~\%$) showed improvements over the individual question approach. The larger Qwen2.5VL reached an $F_1$-score of $40.91~\%$. These results are expected to be improved by using the summary strategy for this model.

Localization experiments examined whether five prompting strategies can locate missing tick marks or labels (Section~\ref{SEC:VisualizationGuidelines} questions 7 and 8). For the horizontal axis, two strategies (introduction question and summary) slightly outperformed the no information rate ($81.63~\%$). Both strategies reached rather poor recall (introduction question: $7.69~\%$, summary: $22.49~\%$), which can be improved by the few-shot approach ($32.54~\%$). The larger 72B version of the Qwen2.5VL ($F_1$-score: $27.59~\%$) reached slightly worse results than the individual question strategy on the Qwen2.5VL-7B  ($F_1$-score: $31.97~\%$). For vertical tick marks and labels, all except the context strategy slightly outperformed the no information rate ($86.20~\%$). The summary strategy reached the best $F_1$-score ($38.34~\%$). For both orientations, the introduction question strategy achieved the lowest recall (horizontal: $7.69~\%$, vertical: $9.45~\%$). The context strategy reached poor precision (horizontal: $19.08~\%$, vertical: $18.21~\%$). The Qwen2.5VL-72B reached an $F_1$-score of $52.42~\%$. Due to these poor results, the individual question strategy was repeated on the proprietary GPT4o model. The $F_1$-scores (horizontal: $35.61~\%$, vertical: $48.20~\%$) outperformed the Qwen2.5VL-7B model. Surprisingly, the vertical axis $F_1$-score does not surpass Qwen2.5VL-72B. For the horizontal axis, a poor recall ($27.81~\%$) was reached. Both orientations achieved accuracies slightly below the no information rate which might be improved by different sampling strategies. As this research focuses on open source models, this point is not elaborated further. The findings indicate improved results for models trained on a larger amount of data. All models struggle more with horizontal tick marks and labels than with vertical ones.

Overall, VLMs can identify missing axis labels in diagrams, with the summary strategy being the most reliable. Compared to this, the detection of missing tick marks and tick labels is more complicated for the Qwen2.5VL model. The recall of the models was improved by a few-shot approach.
%
%
\subsection{Legends}\label{SEC:ResultsLegends}
\begin{table}[t!]
    \centering
    \caption{Results on the identification of legends in $1\,010$ images (Section~\ref{SEC:VisualizationGuidelines} question 11). None of the strategies produced invalid results. The best results are highlighted in bold. No information rate: $52.57~\%$.}
    \label{Table:ResultsLegend}
    \begin{tabular}{|l|r|r|r|r|r|}
    \hline
         Prompting strategy & Precision & Recall &  $F_1$-score & Accuracy & Balanced accuracy\\
         \hline\hline
         Individual question&$92.61~\%$&$\mathbf{99.37~\%}$&$95.87~\%$&$95.94~\%$&$96.11~\%$\\
         Context&$93.49~\%$&$98.96~\%$&$96.15~\%$&$96.24~\%$&$96.37~\%$\\
         Elaborate&$92.79~\%$&$\mathbf{99.37~\%}$&$95.97~\%$&$96.04~\%$&$96.20~\%$\\
         Summary&$\mathbf{94.42~\%}$&$98.96~\%$&$\mathbf{96.64~\%}$&$\mathbf{96.73~\%}$&$\mathbf{96.84~\%}$\\
         Introduction question&$92.72~\%$&$98.33~\%$&$95.44~\%$&$95.54~\%$&$95.68~\%$\\
         \hline
    \end{tabular}
\end{table}

\begin{figure}[t!]
\centering
\includegraphics[width=0.9\textwidth]{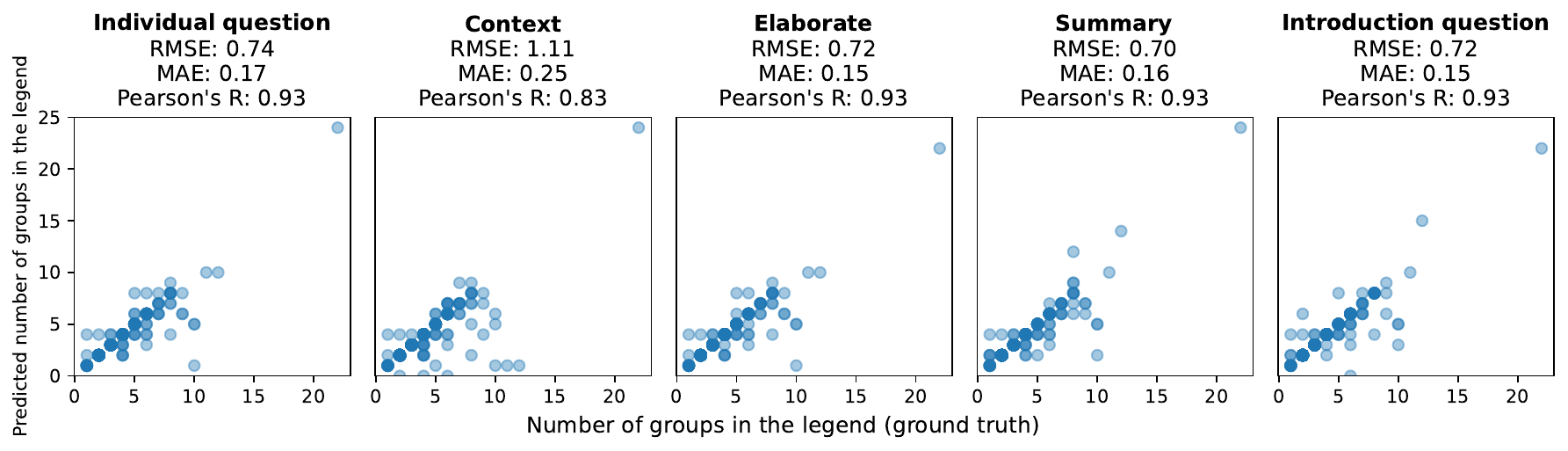}
\caption{Results for the prediction of the number of legend groups (Section~\ref{SEC:VisualizationGuidelines} question 12). These experiments are based on $461$ observations.} \label{Fig:LegendGroups}
\end{figure}

Effective data visualization requires clear, well-structured legends without too many groups (Section~\ref{SEC:VisualizationGuidelines} question 11 and 12). The results for these experiments, which are based on all $1\,010$ images, are summarized in Table~\ref{Table:ResultsLegend}. Results for the 72B model and the few-shot approach are included in the supplementary materials. For $531$ ($52.57~\%$) images, a missing legend was identified. All prompting strategies outperformed the no information rate. The summary strategy achieved the best $F_1$-score ($96.64~\%$) and accuracy ($96.73~\%$). The analysis of legend groups was performed on a set of $461$ images (Section~\ref{SEC:VisualizationGuidelines} question 12). $531$ images without a legend, three with multiple legends, and $15$ with color gradients were excluded. Fig.~\ref{Fig:LegendGroups} shows scatter plots of the ground truth and number of predicted legend groups. Results for the 72B model and the few-shot approach are included in the supplementary materials. The best Pearson's R is $0.93$ (all except the context strategy). The summary strategy reached the best Root Mean Squared Error (RMSE) ($0.70$). The best Mean Absolute Error (MAE) was $0.15$ (elaborate, and introduction question). Qwen2.5VL-72B results outperformed these (RMSE: $0.49$, MAE: $0.10$, Pearson's R: $0.97$).

\subsection{Colors}\label{SEC:ResultsColors}

\begin{figure}[t!]
\centering
\includegraphics[width=0.9\textwidth]{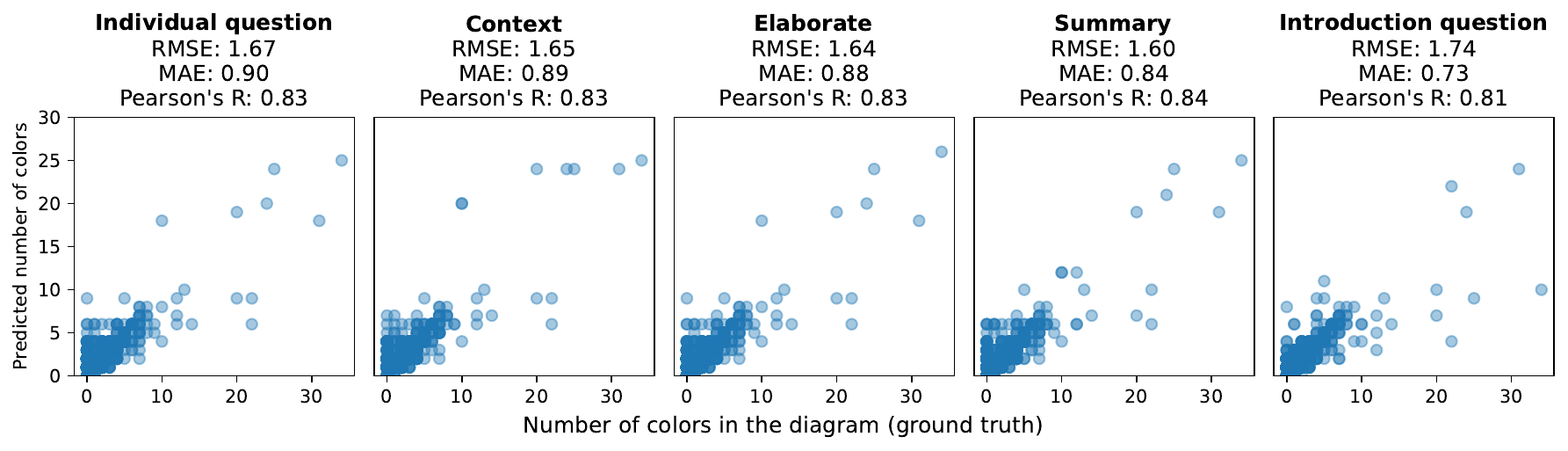}
\caption{Results of the number of colors identified in the diagrams (Section~\ref{SEC:VisualizationGuidelines} question 10). These experiments are based on $915$ observations.} \label{Fig:Colors}
\end{figure}

The number of colors in a diagram is analyzed to avoid confusion from excessive colors (Section~\ref{SEC:VisualizationGuidelines} question 10). These experiments exclude surface diagrams, heatmaps, and Venn diagrams, which often use multiple colors without causing confusion, and images with color gradients leading to $915$ images. Fig.~\ref{Fig:Colors} displays the detected colors vs. manually annotated colors across strategies. Results for the 72B model and the few-shot approach are included in the supplementary materials. All strategies led to plausible results. The summary strategy achieved the best RMSE ($1.60$) and Pearson's R ($0.84$). The best MAE was $0.73$ (introduction question). The Qwen2.5VL-72B reached a decreased MAE of $0.58$ but an increased RMSE ($1.69$) and Pearson's R ($0.82$).

\subsection{Image Quality}
This experiment focuses on detecting compression artifacts and was performed for all $1\,010$ images (Section~\ref{SEC:VisualizationGuidelines} question 13). The results are summarized in Table~\ref{Table:ResultsCompressionartifacts}. Results for the 72B model, the few-shot approach and GPT4o are included in the supplementary materials. 270 images ($26.73~\%$) have compression artifacts (no information rate: $73.27~\%$). None of the strategies outperforms the no information rate. All models reached $F_1$-scores worse than $0.74~\%$ (individual question). The few-shot approach and the 72B model do not show any improvements. Due to the poor results of Qwen2.5VL, the individual question strategy was repeated on the proprietary GPT4o model. This model outperforms Qwen2.5VL ($F_1$-score: $53.28~\%$, precision: $52.52~\%$, recall: $54.07~\%$). The findings suggest that larger models trained on a larger amount of data can improve the results. One explanation for the results is that compression artifacts might be lost during tokenization.

\begin{table}[t!]
    \centering
    \caption{Results of the availability of compression artifacts in $1\,010$ images (Section~\ref{SEC:VisualizationGuidelines} question 13). None of the strategies produced invalid results. The best results are highlighted in bold. No information rate: $73.27~\%$.}
    \label{Table:ResultsCompressionartifacts}
    \begin{tabular}{|l|r|r|r|r|r|}
    \hline
         Prompting strategy & Precision & Recall &  $F_1$-score & Accuracy & Balanced accuracy\\
         \hline\hline
         Individual question&$50.00~\%$&$\mathbf{0.37~\%}$&$\mathbf{0.74~\%}$&$\mathbf{73.27~\%}$&$\mathbf{50.12~\%}$\\
         Context&$0.00~\%$&$0.00~\%$&$0.00~\%$&$73.17~\%$&$49.93~\%$\\
         Elaborate&$16.67~\%$&$\mathbf{0.37~\%}$&$0.72~\%$&$72.87~\%$&$49.85~\%$\\
        Summary&$0.00~\%$&$0.00~\%$&$0.00~\%$&$73.17~\%$&$49.93~\%$\\
         Introduction question&$0.00~\%$&$0.00~\%$&$0.00~\%$&$\mathbf{73.27~\%}$&$50.00~\%$\\
         
         \hline
    \end{tabular}
\end{table}

\subsection{Lines}
\begin{figure}[t!]
\centering
\includegraphics[width=0.9\textwidth]{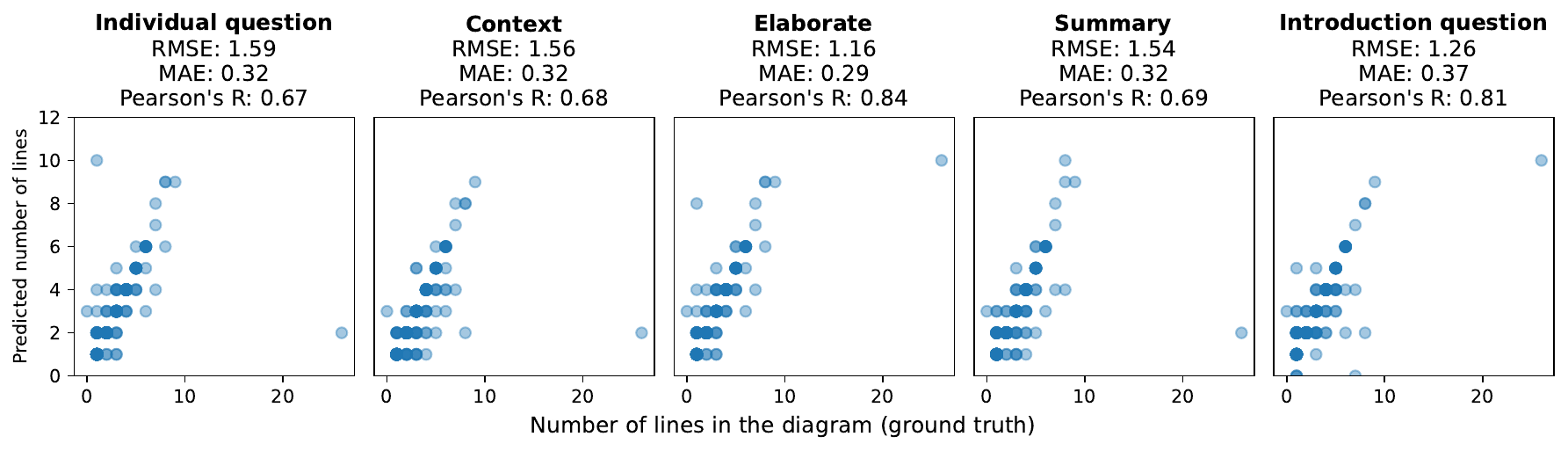}
\caption{Results of the number of lines identified in the diagrams (Section~\ref{SEC:VisualizationGuidelines} question 9). These results are based on 300 observations.} \label{Fig:Lines}
\end{figure}

The number of lines in $300$ line and scatter-line diagrams was analyzed to prevent reader confusion caused by excessive line numbers (Section~\ref{SEC:VisualizationGuidelines} question 9).  Fig.~\ref{Fig:Lines} visualizes the number of VLM-detected lines versus the manual annotations. Results for the 72B model and the few-shot approach are included in the supplementary materials. The scatter plots show a moderate to high correlation for all strategies with the ground truth values. The best Pearson's R was $0.84$ (elaborate strategy). In comparison to the individual question strategy of the Qwen2.5VL-7B, 72B model reached improved results for the RMSE ($1.13$) and the Pearson's R ($0.85$). These results can be utilized to set a threshold for line numbers and provide annotations for diagrams with too many lines.

\section{Discussion and Conclusion}

This work explores the potential of open source VLMs in detecting violations of selected visualization guidelines in scientific diagrams. A subset of 1\,010 diagrams, including twelve diagram types, was sampled from the UB PMC chart dataset. The performance of four general-purpose VLMs and one chart-based VLM in identifying violations of visualization guidelines was evaluated. Five prompting strategies were tested for the best VLM to improve the results. 

The initial, less emphasized question explored which visualization guidelines are ignored in scientific publications. A moderate number of violations were detected, including $16.67~\%$ of the pie charts that used 3D effects, $29.79~\%$ of the diagrams with axes, which lacked axis labels and $22.95~\%$ with missing tick marks or labels. For $52.57~\%$ of the diagrams, no legend was identified and $26.73~\%$ showed compression artifacts. Although these observations are drawn from a relatively small dataset, primarily focused on the medical domain, and are therefore not representative for all research fields, the results still highlight areas for improvement in scientific diagrams. Future work should investigate larger datasets, for example, manually curated sets of the CHART-Info 2024 dataset \cite{Davila2024ChartInfo}, and compare different research areas. Another dataset limitation is the JPG format in PMC, which may introduce compression artifacts. 
In addition, it should be mentioned that visualization guidelines affect diagram types to varying degrees. For example, missing legends may be less critical in Venn diagrams, where sets often directly include labels. Future research is needed to evaluate the connections between visualization guidelines and diagram types. 

The second and main research question examines the capability of VLMs to identify the compliance of diagrams with visualization guidelines. As a preprocessing step, VLMs were asked to identify the diagram type. This step helps as some of the subsequent experiments apply only to specific diagram types; for instance, axis-related analyses are not relevant for pie diagrams. Surprisingly, the chart-specific VLM performed the worst ($F_1$-score: $3.64~\%$) in comparison to general-purpose VLMs as it classified $69.41~\%$ of the diagrams as scatter plots. 
Preliminary experiments with a limited set of chart-specific VLMs showed similar behavior. Further research is required to explore this problem. The best performing VLM was the Qwen2.5VL ($F_1$-score: $82.49~\%$) with default parameters. An experiment was performed to test whether a different search strategy improves the results but it showed no improvement. Further research is needed to assess parameter effects. Qwen2.5VL was used for the remaining experiments, which were performed to identify violations of the selected visualization guidelines. This selection is not exhaustive, allowing for future expansion. Additionally, there is a lack of standardized definitions for some concepts such as compression artifacts, tick marks, or number of valid lines leading to small deviations between the manual annotations, and VLMs. For example, confidence lines were counted as valid lines by manual annotators but often ignored by VLMs.

The experiments performed to identify specific violations of visualization guidelines showed satisfactory results for the detection of 3D effects ($F_1$-score: $98.55~\%$), missing axes labels ($F_1$-score: $76.74~\%$), and missing legends ($F_1$-score: $96.64~\%$). Acceptable performances were reached for counting the legend groups (RMSE: $0.70$), the number of colors (RMSE: $1.60$), and the number of lines (RMSE: $1.16$). The few-shot approach boosted recall on some questions, achieving perfect 3D-effect detection. The 72B model outperformed the smaller version in some cases. However, because of the high resource consumption, the experiments on prompting strategies were carried out with the reduced parameter model. The Qwen2.5VL model showed problems in the detection of tick marks and labels ($F_1$-score: $46.13~\%$), as well as compression artifacts ($F_1$-score: $0.74~\%$). For both experiments, the performance of the proprietary GPT4o model was tested. The GPT4o model showed an increased $F_1$-score ($53.28~\%$) for the detection of compression artifacts and slightly improved $F_1$-scores (horizontal: $35.61~\%$, vertical: $48.20~\%$)  for the detection of missing tick labels and marks. These results might be improved by different prompting strategies. The Qwen2.5VL model was tested with five prompting strategies. In $58.33~\%$ of the experiments, the summary strategy performed best. In the context strategy, errors from previous questions were often carried over, resulting in poor prediction performance. Future work could explore additional prompting strategies and the detection of bounding boxes for specific diagram elements, to enhance user engagement. Additionally, enhanced few-shot methods, combined with varied prompts or fine-tuned models, could further boost performance. For reference it would be interesting to compare the results of the VLMs to those of a pretrained CLIP model.

In summary, this research shows that VLMs can be used to automatically identify a number of potential violations of visualization guidelines in diagrams, such as missing axes labels, missing legends, and 3D effects. 

\begin{credits}
\subsubsection{\ackname}
This work is part of the BMFTR-funded project 
IPPOLIS (Funding code: 16DHBKI050). The work of Louise Bloch was partially funded by a PhD grant from University of Applied Sciences and Arts Dortmund, Dortmund, Germany.
\end{credits}
%
%
%
\bibliographystyle{splncs04}
\bibliography{bibliography}

\end{document}